\documentclass[runningheads]{llncs}
\usepackage[T1]{fontenc}
\usepackage{booktabs}
\usepackage{amsmath}
\usepackage{hyperref}

\usepackage{graphicx}
\usepackage{comment}
\usepackage{subcaption}

\usepackage{color}

\begin{document}
\title{From Regional to Global: Transfer Learning for Atmospheric Transport Emulators}
\titlerunning{Towards global atmospheric transport emulation}
%

\author{Jeff Clark\inst{1}\orcidID{0000-0003-0118-3999} \and
    Elena Fillola\inst{1,2}\orcidID{0000-0003-4706-9833} \and
    Nawid Keshtmand\inst{2}\orcidID{0009-0008-5552-1395} \and
    Raul Santos-Rodriguez\inst{1}\orcidID{0000-0001-9576-3905} \and
    Matthew Rigby\inst{2}\orcidID{0000-0002-2020-9253}}
    \authorrunning{J. Clark et al.}
    %
    \institute{School of Engineering Mathematics and Technology, University of Bristol, UK \and
    School of Chemistry, University of Bristol, UK
    \\
    \email{jeff.clark@bristol.ac.uk}}

\maketitle              
\begin{abstract}
Greenhouse gas emissions estimates can be derived using inverse methods by combining atmospheric concentration observations with chemical transport models. The latter traditionally use physics-driven simulators such as Lagrangian Particle Dispersion Models (LPDMs), which are expensive to run and do not scale well to modern satellites' high resolution data. Previously we developed a performant atmospheric transport emulator that approximates LPDM outputs (``footprints'') over South America $\sim$1{,}000$\times$ faster than the UK Met Office's LPDM. Expanding towards global emulation is not straightforward, as atmospheric transport is regionally heterogeneous.  This paper evaluates spatial transferability capabilities of models across four world regions: South America, East Asia, South Asia, North Africa using both region-specific and multi-region models, and leave-one-region-out experiments.
Regional differences are characterised in the context of input variable and output footprint distributions.
This work builds intuition in cross-region generalisation and transfer learning, aiding regional performance towards efficient global emissions estimates.

\end{abstract}
\section{Introduction}

Accurately quantifying greenhouse gas (GHG) emissions is critical for monitoring progress towards international climate targets such as the Paris Agreement~\cite{paris2015}. Top-down inverse modelling approaches, which infer surface emissions from atmospheric observations, are increasingly important with the availability of high-resolution satellite data such as TROPOMI~\cite{leip2018complete,veefkind2012tropomi}. Inverse methods rely on accurate representations of atmospheric transport to link observed concentrations to upwind emission sources. Lagrangian Particle Dispersion Models (LPDMs), run in a time-reversed mode, are often used to simulate atmospheric transport and generate so-called ``footprints'', which describe the sensitivity of a given observation to surface fluxes. While physically robust, LPDMs are computationally expensive, which limits their applicability to modern satellite datasets. 

Machine learning-based emulation provides a promising alternative. Building on graph-based approaches such as GraphCast~\cite{lam2023learning}, we recently introduced GATES (Graph Neural Network Atmospheric Transport Emulation System)~\cite{fillola2026enabling}, to approximate LPDM outputs at $\sim$1{,}000$\times$ lower computational cost than the UK Met Office's LPDM model. This enables rapid generation of transport footprints, reducing runtime per observation from tens of minutes to less than one second, substantially improving the scalability of inverse modelling systems.

However, extending such approaches from regional to global scales introduces a key challenge: atmospheric transport is highly heterogeneous. Differences in topography, land use, and meteorological regimes lead to footprints with different magnitudes, shapes and characteristics, and data availability creates region-specific biases. As a result, models trained in one region may not generalise effectively elsewhere, limiting the direct applicability of learned emulators. Similar issues have been documented for other applications, such as land-use classification~\cite{zhang2025predicting}.

In this work, we investigate the extent to which atmospheric transport emulators can generalise across regions, and how transfer learning may be applied. Using GATES, we evaluate both region-specific and multi-region models across four regions: South America, East Asia, South Asia, North Africa. Preliminary leave-one-region-out experiments are conducted to provide insight into generalisability to new regions.
We analyse regional differences to better understand the factors governing transfer performance. By characterising regional heterogeneity and transferability, this work contributes towards the development of efficient, scalable, and globally applicable atmospheric transport emulators, enabling improved top-down estimation of greenhouse gas emissions.

\section{Methods}
A GATES model is trained to emulate the outputs from LPDMs, ``footprints'', directly from archived observation-constrained meteorological analyses. Each footprint is a 2-D map which represents, for an atmospheric measurement at a particular location and time, the sensitivity to potential emissions from each grid-cell in the surrounding domain~\cite{fillola2026enabling}. Code is available at the GATES GitHub repository:
\href{https://github.com/GATES-Lab/GATES_LPDM_emulator}{\texttt{GATES\_LPDM\_emulator}}.

\begin{figure}
    \centering
    \includegraphics[width=1\linewidth]{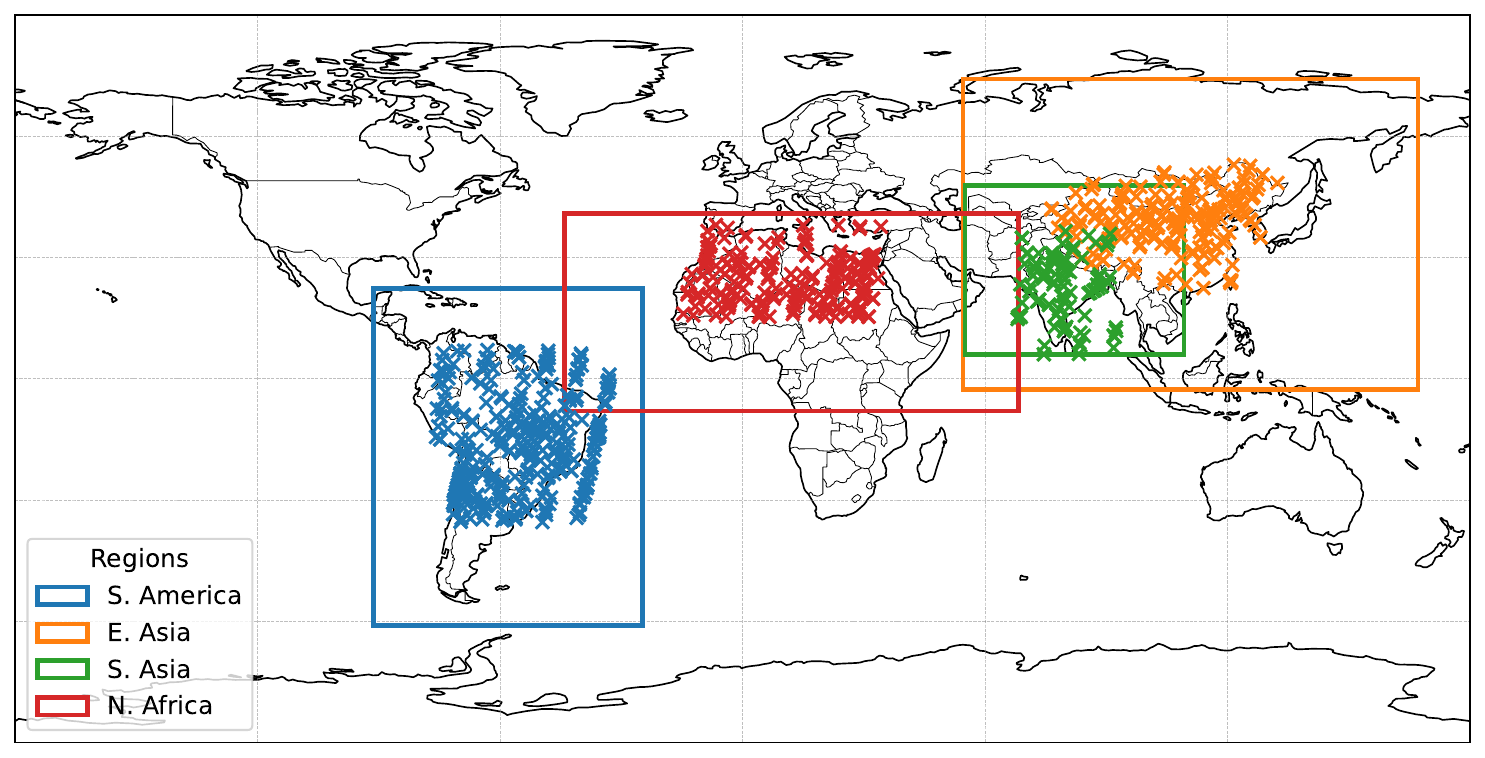}
    \caption{Bounding boxes for the four selected regions, with a subset of test set locations of methane observations (crosses) illustrating spatial coverage.}
    \label{fig:map}
\end{figure}

\smallskip
\noindent
\textbf{Target} We use footprints generated by the UK Met Office Numerical Atmospheric dispersion Model Environment (NAME) LPDM~\cite{jones2007uk} for GOSAT methane observations~\cite{parker2020decade} across four regions: East Asia, South Asia, South America, North Africa (Fig.~\ref{fig:map}), for Jan 2012 to Aug 2015. These regions were chosen due to the availability of training and evaluation data. Footprints are 2-D on a regular $0.352^\circ \times 0.234^\circ$ latitude--longitude grid. They are sparse, with small, exponentially distributed values ($10^{-6}$ to $10^{-1}$ mol mol$^{-1}$ (mol m$^{-2}$ s$^{-1}$)$^{-1}$), and hence are log-transformed during training. 
To reduce sparsity, each data point is cropped to a grid $50 \times 50$ cells centred on the observation location.

\smallskip
\noindent
\textbf{Features} The model meteorological inputs (e.g. wind, pressure) are derived from Met Office Unified Model (UM) global analysis fields, the same that drive the LPDM. Variables are extracted at seven vertical levels (100 m to 18 km) and at three time steps relative to the observation (0, $-6$, and $-12$ hours prior). Static features, including topography and land-use, provide location-specific context. The domain is cropped to that of the footprints, with 160 features in each cell. Further feature details are provided in the original GATES paper~\cite{fillola2026enabling}.

\subsection{Model training}
Data are split temporally into training (2012--2013), validation (Feb--Aug 2015), and testing (2014) sets. To manage dataset size, observations in each dataset are subsampled: South America, East Asia, and North Africa are all sampled at ratios of 3:50:1 (train:val:test), South Asia at 2:25:1. Single-region models were trained, in addition to two multi-region models (MRMs):  
(i) a ``large'' model using all unique sampled datapoints from the four regions, and  
(ii) a ``small'' model trained with proportional subsampling to approximately match the size of individual regional datasets (Table~\ref{table:results}). Leave-one-region-out (LORO) experiments were conducted to assess generalisability to a new region. For this the MRM ``small'' dataset was used, and LORO models were trained using data from all regions except for the target, resulting in four models. Results are presented for the target region. All models were trained for 250 epochs using identical hyperparameters. All models except for the LORO cohort were trained on a single NVIDIA A10 GPU, which took 3--6.5 hours for single-region models, 10 hours for the small multi-region model, and 48 hours for the large multi-region model. LORO training took place later, following some code optimisation and utilising an NVIDIA H100 GPU, took 2--3 hours per model using the Isambard-AI supercomputer~\cite{mcintosh2024isambard}.

\subsection{Performance Evaluation}
Except for the LORO experiments, all models are evaluated on all regions to assess both within-region and cross-region performance:

\subsubsection{Quantitative assessment}
Following previous work~\cite{fillola2026enabling}, emulation performance is measured against ground-truth LPDM-derived footprints using mean squared error (MSE), which captures magnitude accuracy, and intersection over union (IoU), which evaluates spatial overlap and structural similarity, where $b_i$ is the true value and $\hat{b}_i$ is the emulated value.

\begin{enumerate}

    \item Mean Squared Error
    \[
    \mathrm{MSE}(b,\hat{b})
    =
    \frac{1}{n}
    \sum_{i=1}^{n}
    \left( b_i - \hat{b}_i \right)^2
    \]

    \item Intersection Over Union, where $B$ is a binary version of $b$
    \[
    B_{i,j} =
    \begin{cases}
        1 & \text{if } b_{i,j} > 0, \\
        0 & \text{otherwise}
    \end{cases}
    \]

    \[
    \mathrm{IoU}(B,\hat{B})
    =
    \frac{\left| B \cap \hat{B} \right|}
         {\left| B \cup \hat{B} \right|}
    \]

\end{enumerate}

\subsubsection{Qualitative assessment} Visual comparison of predicted and LPDM-derived footprints, alongside regional meteorology, topography, and footprint distributions, to aid interpretation of model performance and transferability.

\begin{table}[t]
\setlength{\tabcolsep}{5pt}
\centering
\caption{Intersection over union ($\uparrow$ is better) / mean squared error ($\mathrm{e}{-7}$, $\downarrow$ is better) test set performance across all models and regions. ``\#'' denotes training samples. MRM = Multi-region model; ``large'' utilises all data, ``small'' utilises proportionally fewer samples equivalent to the single-region models. LORO = leave-one-region-out performance, whereby a model is trained using data from the MRM ``small'' dataset for the three regions except for the target region.}
\label{table:results}
\begin{tabular}{l c c c c c}
\toprule
 &  & \multicolumn{4}{c}{\textbf{Test region}} \\
\cmidrule(lr){3-6}
\textbf{Model} & \textbf{\#} & \textbf{S. America} & \textbf{E. Asia} & \textbf{S. Asia} & \textbf{N. Africa} \\
\midrule
South America & 6920 & 57\% / 9 & 27\% / 20 & 35\% / 31 & 52\% / 6 \\

East Asia & 7263 & 57\% / 53 & 58\% / 10 & 61\% / 22 & 69\% / 24 \\

South Asia & 4542 & 56\% / 19 & 53\% / 21 & 62\% / 20 & 69\% / 9 \\

North Africa & 7811 & 58\% / 14 & 52\% / 14 & 54\% / 23 & 69\% / 4 \\
\midrule
MRM ``large'' & 26449 & 57\% / 8 & 59\% / 10 & 62\% / 21 & 69\% / 4 \\

MRM ``small'' & 7302 & 56\% / 9 & 57\% / 11 & 61\% / 22 & 68\% / 5 \\

LORO & 6065--6849 & 55\% / 12 & 55\% / 15 & 61\% / 19 & 68\% / 7 \\
\bottomrule
\end{tabular}
\end{table}

\begin{figure}[t]
    \centering
    \includegraphics[width=1\linewidth]{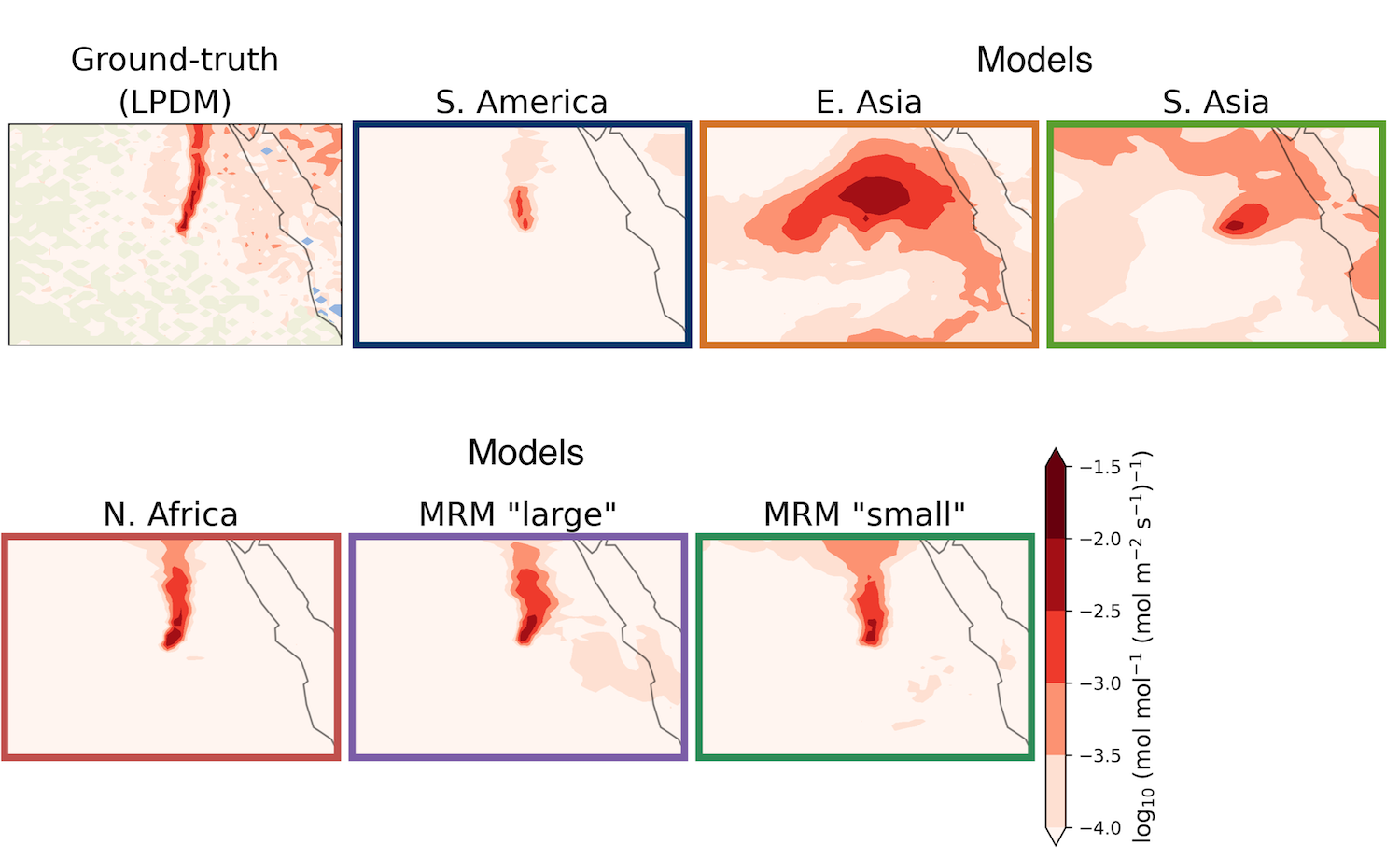}
    \caption{Qualitative footprint emulation performance for a single footprint in North Africa, for each of the single-region and multi-region (MRM) models against the LPDM ground-truth footprint (top left). See Appendix Fig.~\ref{fig:fps_regions} for example emulated footprints from all other regions.}
    \label{fig:fps_NA}
\end{figure}

\section{Results}

\subsection{Quantitative Assessment
}

Best emulation performance for each region was generally achieved either using its native region-specific model, or the large multi-region model (MRM ``large'', Table~\ref{table:results}). For South Asia both the in-region and East Asia-region models appear similarly performant, potentially due to their spatial proximity, wind similarity (Fig.~\ref{fig:topo_wind}) and the low number of native footprints in the training dataset for South Asia (Table~\ref{table:results}). The large multi-region model performed comparably to or better than region-specific models, but at $\sim$2.3$\times$ training cost compared to training all four individual regions (48h vs 21h). The small multi-region model (MRM ``small'') offers reasonably competitive performance across all regions despite only taking 10h to train. MRM ``small'' has the additional advantage of requiring fewer training samples thereby taking significantly less time to generate and facilitate training. The LORO results indicate that reasonable performance can be achieved by utilising a model trained on all regions except for the target, providing reassurance of generalisability to new regions. Performance with the LORO models is generally only slightly worse than the MRM ``small'' configuration despite utilising approximately 25\% less data, and none from the target region. The outlier is South Asia, for which LORO performance is on par or arguably slightly better than models trained using South Asia data. In future work more investigation is required into this, and if similar findings are observed for other regions.

\subsection{Qualitative Assessment
}
Fig.~\ref{fig:fps_NA} demonstrates predictions for a single footprint against the LPDM ground-truth for the North Africa region. Examples for all other regions are available in Appendix Fig.~\ref{fig:fps_regions}. Visually, the North Africa-trained model and the large multi-region model best match the LPDM footprint. The small multi-region model does an admirable job given its limited dataset size. The South America, East Asia, and South Asia-trained models all struggle in different ways on this footprint.

\begin{figure}[t]
    \centering
    \includegraphics[width=1\linewidth]{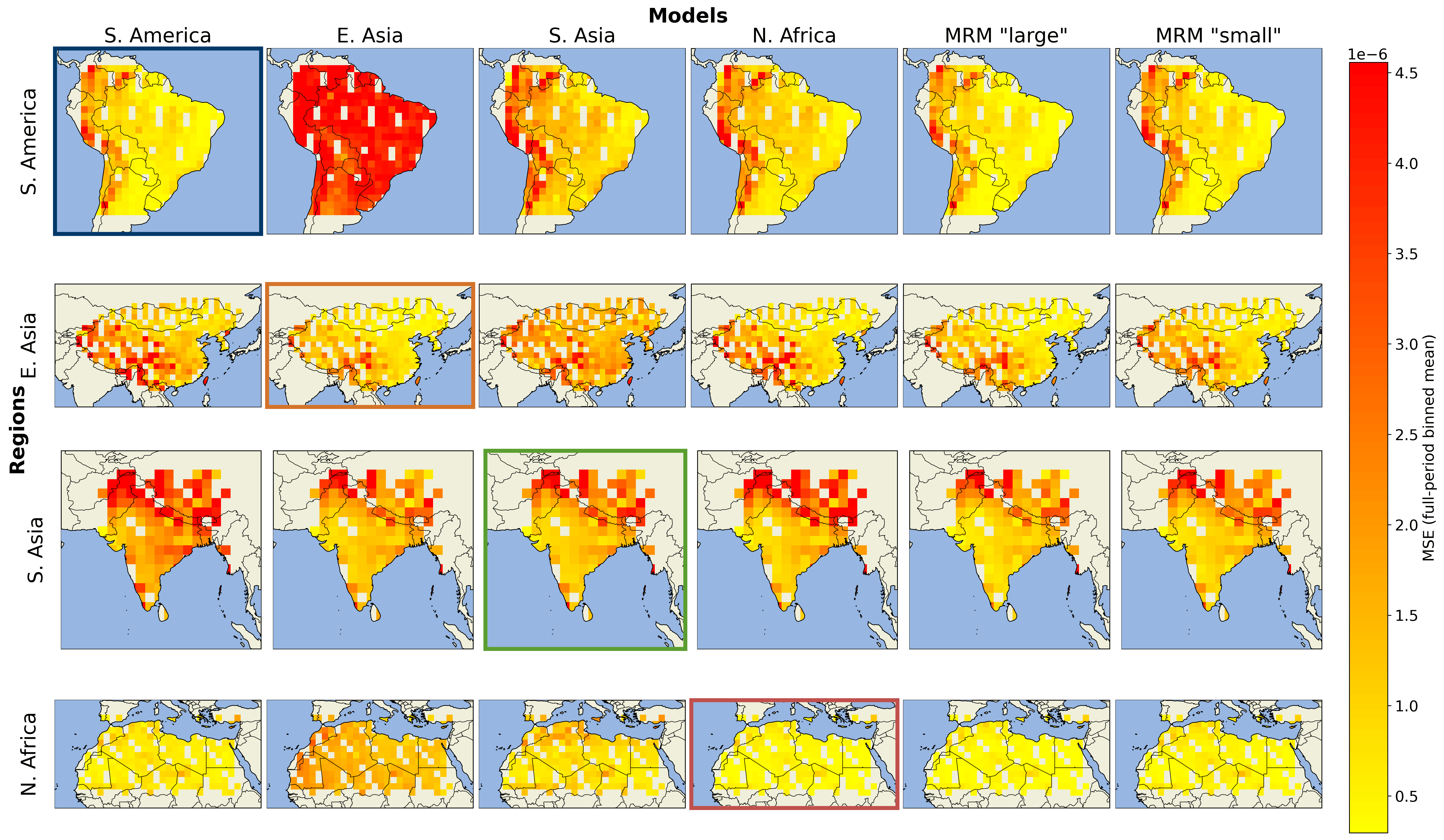}
    \caption{Spatial distribution of mean squared error (MSE) for different model types (columns) evaluated across each test region (rows). The coloured frame in each row highlights the region-specific model for each region.}
    \label{fig:mse_maps_all}
\end{figure}

\begin{figure}[t]
    \centering
    \includegraphics[width=1\linewidth]{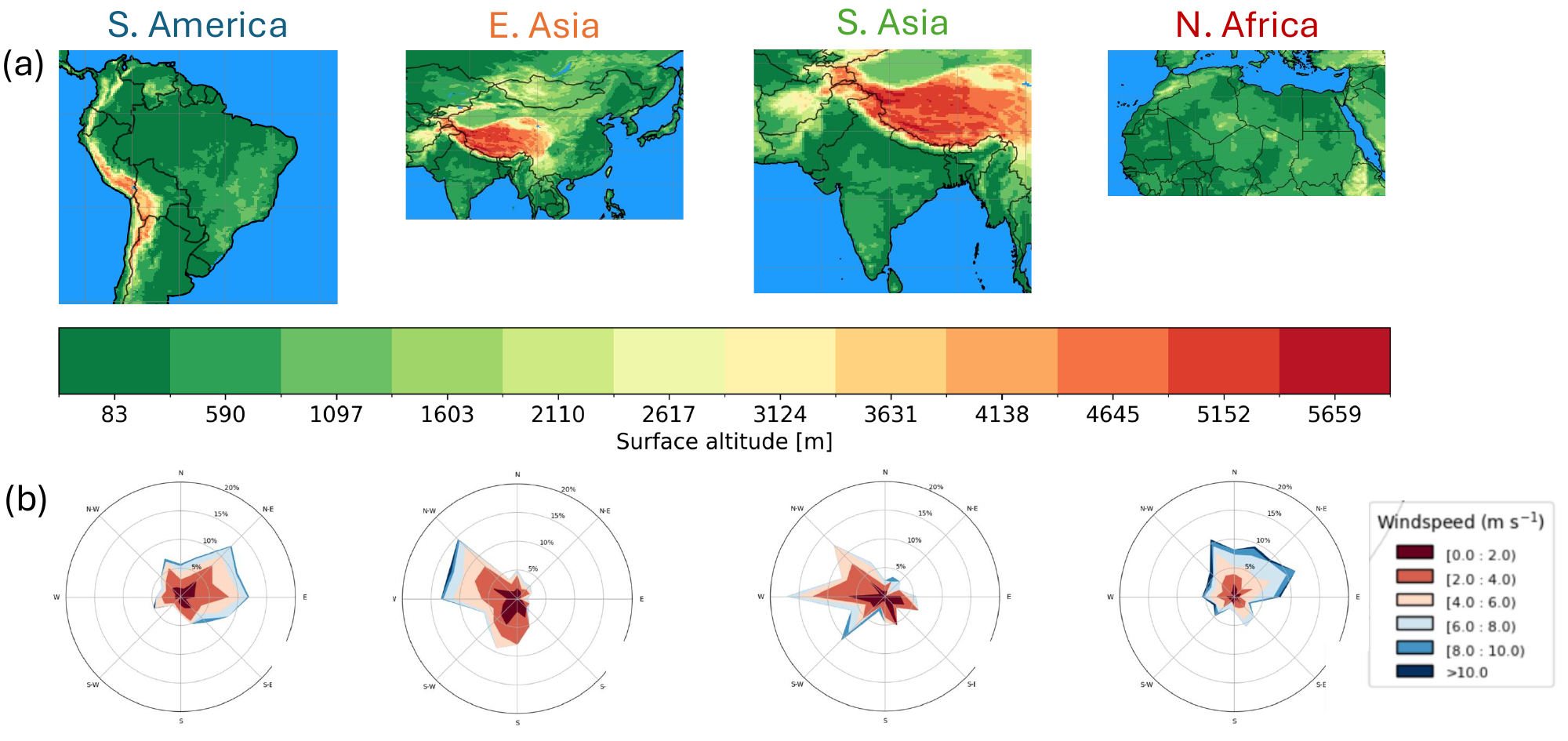}
    \caption{Input feature characterisation for surface (a) topography, and (b) wind roses,  across the four domains.}
    \label{fig:topo_wind}
\end{figure}

The spatial distribution of error across models and regions was evaluated (Fig.~\ref{fig:mse_maps_all}). Aligning with the aggregated results in Table~\ref{table:results}, spatially, models generally achieve the lowest errors within their native training domain and the large multi-region model, while cross-regional evaluation often results in elevated MSE and more spatially heterogeneous error patterns. The stable performance of multi-region models across all domains suggests improved generalisation under geographically diverse training data. Errors are particularly pronounced using the East Asia model to predict on South America, potentially caused by the large difference in wind inputs (Fig.~\ref{fig:topo_wind}). All models consistently struggled with high errors within mountainous regions (Fig.~\ref{fig:mse_maps_all} and Fig.~\ref{fig:topo_wind}), which are noted to exhibit complex atmospheric transport regimes~\cite{miller2015biases}.

\begin{figure}
    \centering
    \includegraphics[width=0.8\linewidth]{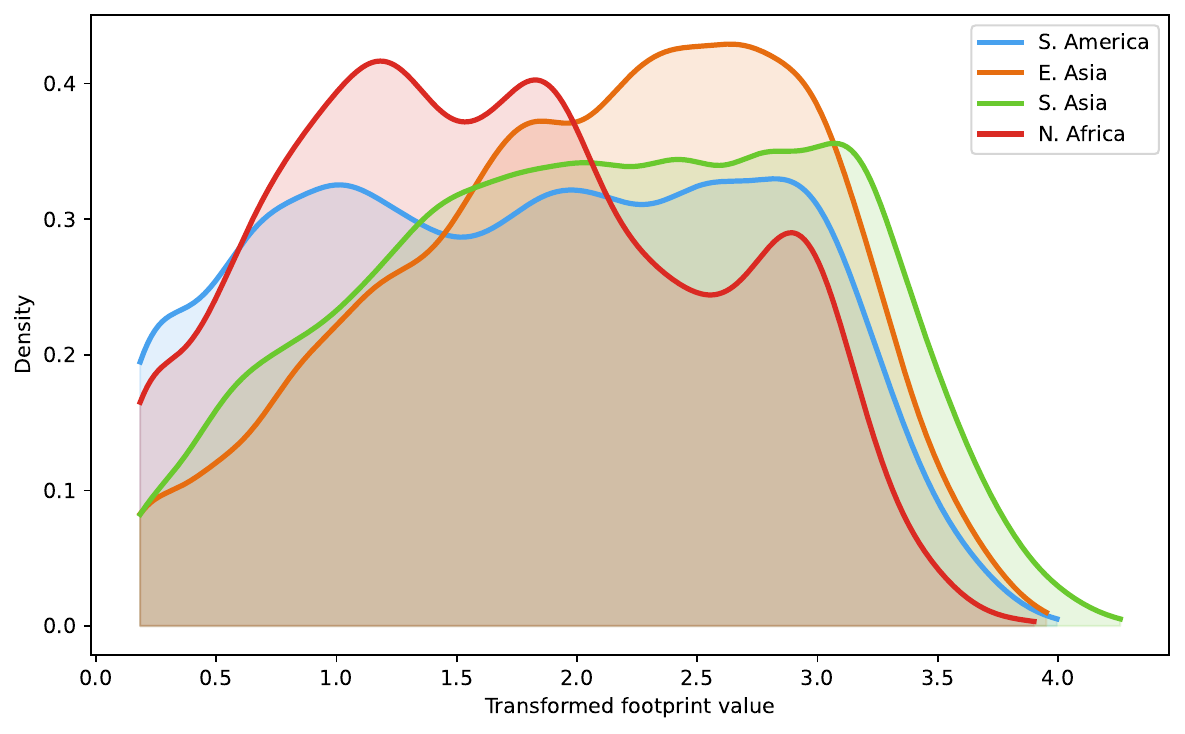}
    \caption{Distribution of non-zero footprint values 
Kernel Density Estimate (KDE) of the flattened footprint values in the transformed space log10(fp) + 5}.
    \label{fig:fps_distributions}
\end{figure}

These results demonstrate that atmospheric transport emulation is region-dependent: single-region models perform best within-region, while transfer of single-region models across distinct regions degrades performance, likely due to differences in wind regimes, topography, and footprint distributions (Fig.~\ref{fig:topo_wind}, Appendix Fig.~\ref{fig:fps_distributions}).
In contrast, multi-region models achieve more consistent performance across regions, indicating improved generalisation from geographically diverse training data. Notably, the strong performance of the smaller multi-region model and leave-one-region-out experiments suggests that careful sampling strategies may enable robust global emulators without substantially increasing training data requirements.

\section{Conclusion}
This work investigated how transferable atmospheric transport emulators are across global regions. While region-specific models generally achieved the best within-domain performance, cross-regional evaluation demonstrated that atmospheric transport is domain-dependent, with variations in topography, meteorology, and footprint distributions impacting generalisation. Multi-region training improved robustness across domains, highlighting the value of geographically diverse datasets for scalable transport emulation. The large multi-region model achieved performance comparable to or exceeding region-specific models across all test domains, demonstrating the feasibility of developing globally applicable atmospheric transport emulators. The strong performance of the smaller multi-region model suggests that efficient sampling strategies may reduce training costs while maintaining generalisation capability. Preliminary leave-one-region-out experiments suggest that performance using models trained on small amounts of data from multiple regions holds for deployment on unseen regions.

More broadly, these findings highlight challenges of domain shift in geophysical machine learning, where models must generalise across heterogeneous spatial regimes. Future work will investigate improved sampling approaches, applicability maps, additional regions, the development of practical guidelines for best approaches to train an emulator for a new region, and the trade-off between locality and coverage by exploring finer-grained regional models (e.g., country- or sub-region-level), to determine where increased localisation ceases to yield meaningful performance gains over broader, more scalable models. Future work may also consider alternative metrics to better discern relative model performance, use of structural similarity index measure (SSIM)~\cite{wang2004image} to capture structural similarity, statistical testing, and the explicit use of transfer learning strategies to adapt an existing model to a new domain should be considered, such as fine-tuning and domain adaptation.

Overall, this work provides insight into the spatial transferability capabilities of emulation models for atmospheric transport modelling across geographical regions, supporting development of efficient global-scale inverse modelling systems for greenhouse gas emissions estimation.

\begin{credits}
\subsubsection{\ackname} This study was funded by UKRI NERC grant NE/Z504294/1. Compute for all research except for LORO experiments was provided by the University of Bristol using Oracle Cloud credits. The authors acknowledge the use of resources provided by the Isambard-AI National AI Research Resource (AIRR) for the LORO experiments. Isambard-AI is operated by the University of Bristol and is funded by the UK Government’s Department for Science, Innovation and Technology (DSIT) via UK Research and Innovation; and the Science and Technology Facilities Council [ST/AIRR/I-A-I/1023].

\subsubsection{\discintname}
The authors have no competing interests to declare that are
relevant to the content of this article.
\end{credits}

%
\bibliographystyle{splncs04}
\bibliography{ref}

\section{Appendix}

\subsection{Additional figures}

\begin{figure}
    \centering

    \begin{subfigure}{\linewidth}
        \centering
        \includegraphics[width=\linewidth]{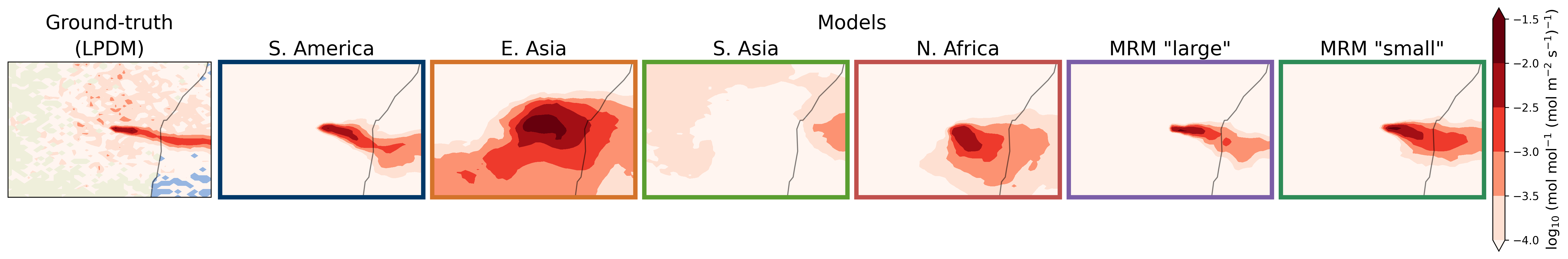}
        \caption{South America footprint}
        \label{fig:fps_SA}
    \end{subfigure}

    \vspace{0.5em}

    \begin{subfigure}{\linewidth}
        \centering
        \includegraphics[width=\linewidth]{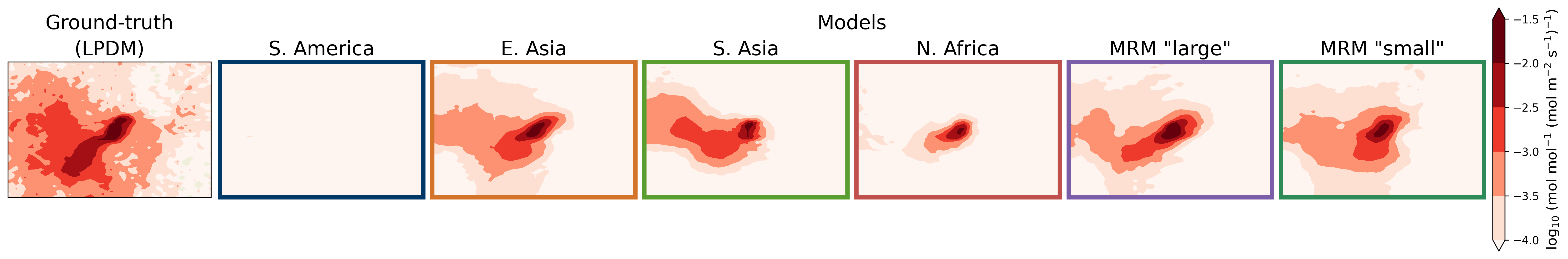}
        \caption{East Asia footprint}
        \label{fig:fps_China}
    \end{subfigure}

    \vspace{0.5em}

    \begin{subfigure}{\linewidth}
        \centering
        \includegraphics[width=\linewidth]{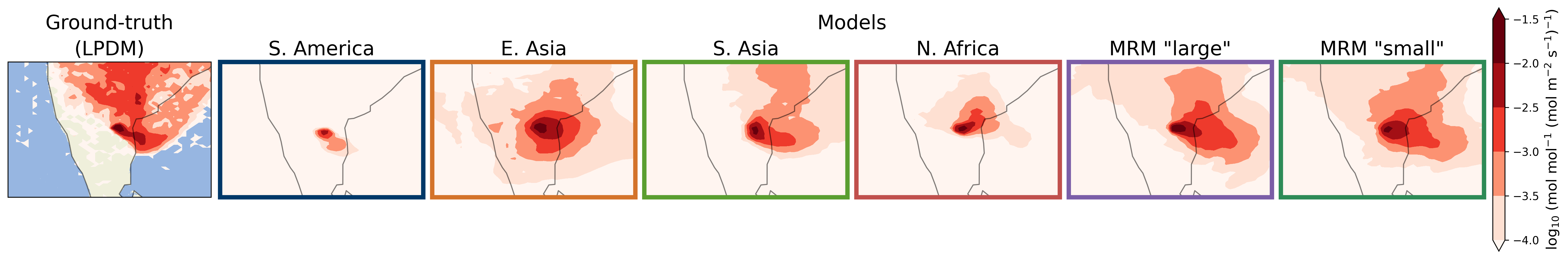}
        \caption{South Asia footprint}
        \label{fig:fps_India}
    \end{subfigure}

    \caption{Comparison of single-region and multi-region (MRM) model outputs for a single footprint from South America (a), East Asia (b), and South Asia (c), against the LPDM ground-truth footprint (far left). For a North Africa footprint see Fig.~\ref{fig:fps_NA}.}
    \label{fig:fps_regions}
\end{figure}

\end{document}